\documentclass[conference]{IEEEtran}
\IEEEoverridecommandlockouts

\usepackage{cite}
\usepackage{amsmath,amssymb,amsfonts}
\usepackage{algorithmic}
\usepackage{graphicx}
\usepackage{booktabs}
\usepackage{textcomp}
\usepackage{xcolor}
\def\BibTeX{{\rm B\kern-.05em{\sc i\kern-.025em b}\kern-.08em
    T\kern-.1667em\lower.7ex\hbox{E}\kern-.125emX}}
\begin{document}

\title{Evaluating Vision Language Model Adaptations for Radiology Report Generation in Low-Resource Languages}

\author{
    \IEEEauthorblockN{Marco Salmè\IEEEauthorrefmark{1},
                      Rosa Sicilia\IEEEauthorrefmark{1},
                      Paolo Soda\IEEEauthorrefmark{1}\IEEEauthorrefmark{2},
                      and
                      Valerio Guarrasi\IEEEauthorrefmark{1}}
    \IEEEauthorblockA{\IEEEauthorrefmark{1}Research Unit of Computer Systems and Bioinformatics, Department of Engineering, \\
    Università Campus Bio-Medico di Roma, Rome, Italy \\
    Email: marco.salme@unicampus.it, r.sicilia@unicampus.it, p.soda@unicampus.it, valerio.guarrasi@unicampus.it}
    \IEEEauthorblockA{\IEEEauthorrefmark{2}Department of Diagnostics and Intervention, Radiation Physics, Biomedical Engineering, Umeå University, Umeå, Sweden \\
    Email: paolo.soda@umu.se}
}

\maketitle

\begin{abstract}
The integration of artificial intelligence in healthcare has opened new horizons for improving medical diagnostics and patient care. 
However, challenges persist in developing systems capable of generating accurate and contextually relevant radiology reports, particularly in low-resource languages.
In this study, we present a comprehensive benchmark to evaluate the performance of instruction-tuned Vision-Language Models (VLMs) in the specialized task of radiology report generation across three low-resource languages: Italian, German, and Spanish.
Employing the LLaVA architectural framework, we conducted a systematic evaluation of pre-trained models utilizing general datasets, domain-specific datasets, and low-resource language-specific datasets. 
In light of the unavailability of models that possess prior knowledge of both the medical domain and low-resource languages, we analyzed various adaptations to determine the most effective approach for these contexts.
The results revealed that language-specific models substantially outperformed both general and domain-specific models in generating radiology reports, emphasizing the critical role of linguistic adaptation. 
Additionally, models fine-tuned with medical terminology exhibited enhanced performance across all languages compared to models with generic knowledge, highlighting the importance of domain-specific training.
We also explored the influence of the temperature parameter on the coherence of report generation, providing insights for optimal model settings. 
Our findings highlight the importance of tailored language and domain-specific training for improving the quality and accuracy of radiological reports in multilingual settings.
This research not only advances our understanding of VLMs adaptability in healthcare but also points to significant avenues for future investigations into model tuning and language-specific adaptations.
\end{abstract}


\section{Introduction}
\label{intro}
Foundation Models (FMs)~\cite{bommasani2021opportunities} represent a groundbreaking advancement in artificial intelligence, bringing significant improvements across numerous disciplines, including medicine~\cite{zhang2024challenges}.
FMs are a class of large-scale models trained on different and extensive datasets, enabling them to generalize across a wide range of tasks and domains.
In medicine, FMs demonstrate their potential through multimodal capabilities, processing and integrating various types of data, such as textual records, medical images, and structured patient information. 
These models, which simultaneously combine visual and textual data analysis, are commonly referred to as Vision-Language Models (VLMs)~\cite{van2024large}. 
This multimodal approach enables the integration of diverse patient information, leading to a comprehensive understanding of patient's health~\cite{guarrasi2024systematic,guarrasi2024multimodal,guarrasi2023multi,rofena2024deep,ruffini2024multi}. 
By synthesizing textual records, medical images, and structured data, these models support healthcare professionals in diagnosing, treating, and managing conditions through applications like diagnostic assistance, treatment planning, and automated medical report generation.
Training FMs typically involves three key phases: pre-training, instruction-tuning, and fine-tuning, each designed to progressively refine the model's capabilities for generalization and task-specific performance.
In the pre-training phase, the model is trained on large, different datasets, enabling it to acquire a broad understanding of both image and text data~\cite{yang2024vision}. 
This is followed by the instruction tuning phase, where the model learns to follow textual instructions, enhancing its ability to generate responses and process tasks based on visual and textual inputs~\cite{mishra2021cross,wei2021finetuned,sanh2021multitask}. 
Finally, during the fine-tuning phase, the model is adapted to a specific downstream task by training on specialized, domain-specific datasets~\cite{yu2024visual}.
Each stage of this training paradigm is crucial, enabling the models to achieve high performance and adaptability across various applications, including those in complex fields like healthcare.
In the context of fine-tuning, adaptation refers to tailoring the model to better perform on specific domains.
We can distinguish between two types of adaptation: language adaptation, which focuses on adapting the model to new languages, and domain adaptation, which targets adapting the model to specific knowledge areas or fields.

Language adaptation is particularly valuable for extending the model's capabilities to underrepresented languages, enabling inclusivity and broader accessibility.
Nevertheless, it is particularly challenging because of VLM's English-centric training foundations~\cite{nguyen2024multilingual, wang2023adapting}. 
The predominant bias towards English-language training data creates significant limitations for models attempting to operate effectively across different linguistic landscapes~\cite{zhang2023don}. 
This issue is particularly pronounced for low-resource languages~\cite{magueresse2020low}, which suffer from systemic under-representation in both pre-training corpora and annotated datasets.
Several state-of-the-art methodologies have been proposed to address these limitations~\cite{muennighoff2022crosslingual,chen2023monolingual,shaham2024multilingual}.
Muennighoff et al.~\cite{muennighoff2022crosslingual} investigated the zero-shot generalization capabilities of LLMs, emphasizing their ability to generalize effectively across various languages. 
However, their analysis focused solely on generalization to languages encountered during pre-training, without exploring the potential for generalization to languages introduced exclusively during fine-tuning.
Chen et al.~\cite{chen2023monolingual} examined the effects of monolingual and multilingual instruction tuning, showing that in a resource-constrained environment, multilingual tuning offers significant advantages over monolingual tuning.
Similarly, Shaham et al.~\cite{shaham2024multilingual} analyzed monolingual and multilingual instruction tuning, showing that models trained on multilingual datasets achieve superior performance to monolingual models while requiring significantly fewer examples per language.
Despite these contributions, a comprehensive analysis of the impact of various instruction tuning approaches on subsequent downstream task fine-tuning remains lacking.

Domain adaptation ensures the model performs optimally in specialized applications, such as legal analysis or medical diagnosis, where domain-specific knowledge is essential.
The unique terminologies and clinical contexts specific to healthcare, as well as data scarcity, make it critical. 
To further improve the performance of VLMs in medical downstream tasks, Supervised Fine-Tuning (SFT) is typically conducted using datasets that are specifically designed for those tasks~\cite{chen2023meditron,li2024llava,hyland2023maira}.
Chen et al.~\cite{chen2023meditron} fine-tuned MEDITRON on MedQA, PubMedQA, and MedMCQA datasets to enhance its performance in medical question answering. 
Similarly, Li et al.~\cite{li2024llava} improved the performance of LLaVA-Med in medical Visual Question Answering (VQA) by training it on PathVQA, SLAKE, and VQA-RAD datasets. 
Following a similar methodology, Chaves et al.~\cite{chaves2024towards} developed LLaVA-Rad, specifically enhancing its performance in Radiology Report Generation (RRG).
Within the context of RRG, Hyland et al.~\cite{hyland2023maira} fine-tuned Vicuna~\cite{chiang2023vicuna} using large-scale image-report pairs to develop the MAIRA-1 model, utilizing RAD-DINO~\cite{perez2024rad} as the encoder.
A significant limitation of these approaches is the use of LLMs that have not undergone comprehensive pre-training on a broad medical domain corpus.
As the adaptation of VLMs to low-resource languages and specialized domains progresses, four challenges persist that underline the need for ongoing research and development.
First, despite recent works suggesting various approaches~\cite{yong2022bloom+,yang2023bigtranslate}, it is currently still unclear how to dynamically and efficiently extend languages for VLMs. 
Second, the limited number of datasets available in languages other than English significantly restricts the possibility to evaluate the performance of VLMs on other languages.
Third, with reference to medical domain adaptation, Nicolson et al.~\cite{nicolson2023improving} explored the benefits of exploiting a warm start to improve the RRG task, but no work has currently investigated this for languages other than English and for models as large as LLMs.
The fourth, and main challenge, is the absence of LLMs that combine deep expertise in the medical domain with proficiency in languages other than English, primarily due to the lack of sufficient data. 
This highlights the need to explore approaches to address this gap.

In this work, we present a comprehensive benchmark to investigate whether starting from an instruction-tuned model, with domain-specific or language-specific knowledge, offers advantages for downstream tasks. Specifically, we focus on evaluating these benefits in the context of RRG.
Our benchmark encompasses three underrepresented languages, i.e., Italian, German, and Spanish, which are significantly less represented compared to English in existing datasets. 
Extensive experiments were conducted by testing several LLMs pre-trained on different datasets.

In summary, our contributions are: 
\begin{itemize}
    \item We establish a baseline for RRG in English, a high-resource language. 
    This includes a comparative analysis between a generalist model and one pre-trained with domain-specific medical knowledge, highlighting the impact of domain expertise on performance.
    \item We assess the effect of medical instruction tuning, primarily conducted in English, on model performance for RRG in low-resource languages. 
    \item We evaluate the contribution of language-specific instruction tuning for RRG in low-resource languages. 
    \item Given the absence of VLMs that have undergone both medical and language-specific instruction tuning, we conduct a comparative study to determine which approach is more effective for RRG. 
    \item We perform a comparative analysis of language-specific instruction tuning for RRG across three low-resource languages. 
    \item We analyze the significance of the temperature parameter in generating radiology reports, exploring how its adjustment impacts coherence and randomness, and provide guidelines for optimal tuning based on task and adaptation needs.
\end{itemize}

The remainder of this paper is organized as follows: section~\ref{methods} provides a detailed explanation of the approach adopted to establish the framework for comparing models; in section~\ref{ExpSetup} we outline the experimental setup, which includes the dataset description, training procedures, and evaluation metrics; subsequently, in section~\ref{results} we present and discuss the results of our experiments; finally, section~\ref{conclusions} summarizes our findings and provides concluding remarks.

\section{Methods}
\label{methods}
In this section, we describe the methodological framework designed to evaluate the effectiveness and adaptability of VLMs in generating medical reports across underrepresented languages. 
Our investigation involved a series of experiments aimed at comparing the performance of various LLMs in adapting to the RRG task, between three low-resource languages, i.e., Italian, German, and Spanish. 
This comprehensive approach establishes a robust benchmark for assessing the impact of domain-specific fine-tuning on the capabilities of VLMs in multilingual medical applications. 
It should be noted that designating Italian, German, and Spanish as low-resource languages may not be entirely precise; however, within the context of this study, they are classified as such due to the substantially lower volume of textual data available compared to English.

\begin{figure}
    \centering
    \includegraphics[width=0.45\textwidth]{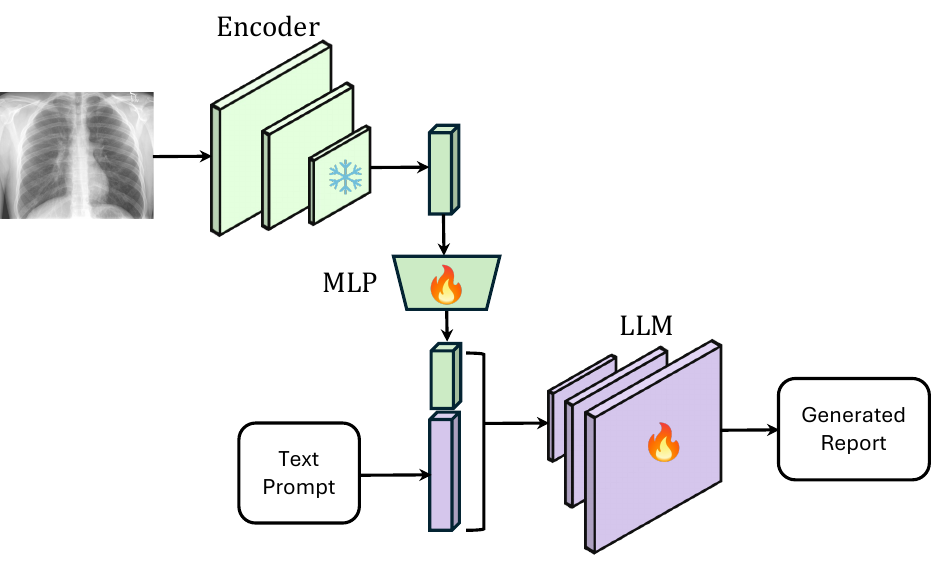}
    \caption{The used architectural framework.The instruction-tuned VLM generates radiology reports based on a fixed input prompt: ``Provide the findings of the following radiology image.''.}
    \label{fig:1}
\end{figure}

Fig.\ref{fig:1} illustrates the architecture used in our study, which employs the LLaVA framework~\cite{li2024llava}, integrating a frozen image encoder with a LLM serving as the decoder.
The first block in the figure depicts the image encoder, specifically the MedSAM encoder~\cite{ma2024segment}, selected for its robust performance in medical semantic analysis. 
This encoder processes input medical images to extract detailed feature representations. 
The frozen configuration of the MedSAM encoder ensures that it functions solely as a feature extractor without undergoing modifications during the training process.
The small block in the middle highlights the multimodal adapter, implemented as a Multi-Layer Perceptron (MLP). This component establishes a connection between the image features and the word embedding space by projecting the extracted image features into a format compatible with the LLM decoder. 
The adapter serves as the intermediary that enables the seamless integration of visual and textual modalities.
The third block is the LLM decoder, which is fine-tuned using LoRA (Low-Rank Adaptation)~\cite{hu2021lora} on radiology reports. 
This step aims to adapt the decoder to medical language and task-specific nuances while preserving computational efficiency. 
Fine-tuning is confined to the decoder and the multimodal adapter, enabling a focused evaluation of the LLM's adaptability and its specific capability to generate radiological reports.

\begin{figure*}
    \centering
    \includegraphics[width=0.8\textwidth]{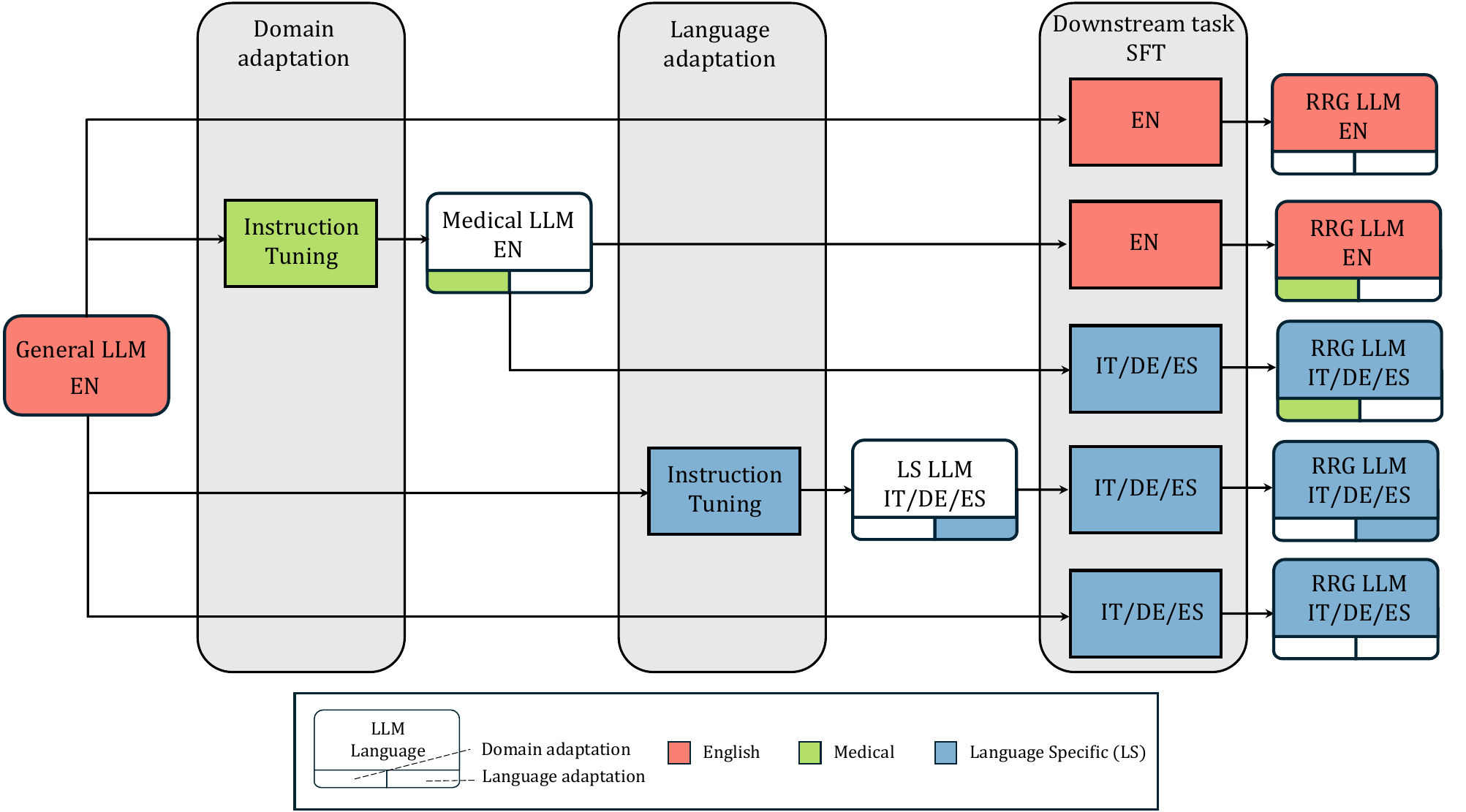}
    \caption{Schematic representation of the methodological approach.}
    \label{fig:2}
\end{figure*}

Fig.\ref{fig:2} presents the methodological approach followed in our experiments. 
From left to right, the first two gray panels illustrate adaptations made at the instruction-tuning level: the first focuses on the medical domain, while the second addresses linguistic adaptation. 
The third panel depicts the systematic evaluation of the decoder component, performed using five distinct configurations of LLMs, each fine-tuned on radiology reports in various languages, as indicated in the corresponding rectangular blocks.
To establish a baseline, two initial experiments were conducted using configurations focused on English-language medical reports, as illustrated in the first two rows of Fig.\ref{fig:2}. 
The first configuration utilized a general-purpose LLM (General LLM) pre-trained predominantly on English data, with minimal exposure to other languages and no specialized medical knowledge. 
This setup was designed to evaluate the generalizability of non-specialist LLMs in generating English medical reports. 
The second configuration involved enhancing the general-purpose LLM, which will be referred to as the Medical LLM. 
This enhancement was achieved by tuning the model on English medical texts to incorporate domain-specific knowledge. 
This configuration was compared against the general-purpose model to quantitatively assess the benefits of specialized fine-tuning for medical applications.
The remaining three experiments, corresponding to the last three rows of Fig.\ref{fig:2}, investigated the capabilities and limitations of LLMs in generating medical reports in Italian, German, and Spanish. 
The third experiment assessed the performance of the limited multilingual Medical LLM fine-tuned on radiology reports written in these target languages. 
This configuration aimed to evaluate the advantages of medical expertise enhancement while highlighting the constraints posed by limited linguistic adaptation.
The fourth experiment examined Language-Specific LLMs tailored for Italian, German, and Spanish. 
These models were derived from the general-purpose LLM, followed by instruction-tuning on general corpora specific to each language without exposure to medical data. 
Subsequent fine-tuning adapted these Language-Specific LLMs to generate medical reports specifically written in the corresponding languages, offering insights into the role of linguistic adaptation in the absence of domain-specific pre-training.
Finally, the fifth experiment evaluated the general-purpose LLM’s ability to generalize across languages and domains. 
This model, pre-trained predominantly in English and without specialized knowledge of the medical domain or the target languages, was directly assessed for its performance in generating medical reports in Italian, German, and Spanish. 

The LLaVA framework provides the foundation for these experiments, offering a structured approach to adapt FMs to complex, language-sensitive tasks such as RRG. 
The General LLM, used as backbone for this study, is Mistral-Instruct 7B~\cite{jiang2023mistral}, chosen for its efficiency and superior performance relative to other state-of-the-art models with the same number of parameters. 
The Instruct variant of Mistral 7-B was specifically selected for its enhanced capability to follow instructions and perform specialized tasks, such as medical report generation.
For the Medical LLM, we employed BioMistral~\cite{labrak2024biomistral}, a fine-tuned version of Mistral-Instruct specifically adapted to medical texts, primarily in English, sourced from PubMed~\cite{mcentyre2001pubmed}. 
For the Language-Specific LLMs, we utilized Maestrale~\cite{huggingfaceMaestrale}, LeoLM~\cite{leolm2023}, and Occiglot-es~\cite{huggingfaceOcciglot}, fine-tuned versions of Mistral-Instruct tailored to Italian, German, and Spanish datasets, respectively.
Tab. \ref{tab:model_comparison} provides a comparison of various LLMs, highlighting their parameter counts, pre-training data, language coverage, domain, and the size of their instruction tuning datasets. 
However, one significant challenge in summarizing these models lies in the heterogeneity of the units used to measure the size of their instruction tuning datasets. 
This ranges from the number of tokens, to the size in megabytes, the type of interaction such as conversations, and even percentages specifying the linguistic composition of the data. 
Additionally, the sources of data are not always transparent, which can further complicate efforts to standardize comparisons. 

\begin{table*}[ht]
\caption{Summary of key characteristics of the models used in this study.}
\label{tab:model_comparison}
\centering
\begin{tabular}{l|l|l|l|l|l}
\toprule
\textbf{Model} & \textbf{\# Parameters} & \textbf{Pre-training Data} & \textbf{Language Coverage} & \textbf{Domain} & \textbf{Instruction Tuning Data} \\
\midrule
Mistral Instruct\cite{jiang2023mistral}    & 7B                       & N/A                                 & EN, Limited Multilingual          & General                  & N/A                                    \\ \hline
Biomistral~\cite{labrak2024biomistral}    & 7B                      & N/A                              & EN, Limited Multilingual  & Medicine                 & 3B Tokens                                    \\ \hline
LeoLM~\cite{leolm2023}    & 7B                       & 65B Tokens                             & EN, DE               & General            & 7 MegaByte                                       \\ \hline
Occiglot-es~\cite{huggingfaceOcciglot}    & 7B                      & 52\% Spanish                             & EN, ES & General    & 160M Tokens                                     \\ \hline
Maestrale~\cite{huggingfaceMaestrale}           & 7B                      & N/A                              & EN, IT          & General                  & 1.7M conversations                                       \\
\bottomrule
\end{tabular}
\end{table*}

\section{Experimental Setup}
\label{ExpSetup}
This section outlines the experimental setup of our study, detailing the dataset configuration, model hyperparameters, and evaluation protocols used to assess the quality of the generated medical reports.

\subsection{Materials}
We trained and evaluated our models on the IU-Xray dataset~\cite{demner2016preparing}, a well-established resource in the field of medical image analysis. 
The dataset consists of 7,470 chest X-ray images paired with 3,955 radiological reports, making it a robust foundation for training and evaluation. 
We selected this dataset to explore the efficient adaptation of FMs using a relatively small dataset.
The dataset includes frontal and lateral radiological images for most reports. Consistent with prior studies~\cite{chen2022cross, chen2020generating, liu2021exploring}, we excluded samples without a ``Findings'' section, as this section provides the essential ground truth for SFT in report generation. 
We used a standard dataset split of 70\% for training, 10\% for validation, and 20\% for testing, ensuring that the same patients were kept within a single split.
Additionally, we utilized Google Translate APIs~\cite{google_translate} to translate the reports from English into Italian, German, and Spanish, generating multilingual versions of the dataset.
For image preprocessing we strictly adhered  to MedSAM’s protocol~\cite{ma2024segment}, with the original image dimensions of $2496 \times 2048$ resized to $1024 \times 1024$ to ensure alignment with the requirements of the frozen encoder. 
Before resizing, a center crop was applied, taking the smaller dimension as reference, to create square images and preserve human anatomical proportions.

\subsection{Training details}
We followed the LLaVA training procedure~\cite{li2024llava}, which is based on two distinct stages. 
In the first stage, the MLP was trained as an adapter for one epoch, with both the encoder and decoder kept frozen. 
This process was designed to align the embedding space of image features with that of words, ensuring compatibility between the modalities. 
The second stage involved an efficient fine-tuning using LoRA for 5 epochs, optimizing both the MLP and the LLM with a standard auto-regressive language modeling loss~\cite{graves2013generating}. 
The training process utilized a cosine learning rate scheduler with a warm-up of 0.03, a learning rate of $2 \times 10^{-5}$, and a batch size of 16. Validation metrics were monitored throughout the training, and the model checkpoint with the lowest validation loss was selected for evaluation.
All experiments were conducted using an NVIDIA A100fat GPU. 
The first stage of training for a single experiment required approximately 2 GPU hours, while the LoRA fine-tuning stage took around 12 GPU hours to complete.

\subsection{Evalutation Metrics}
To evaluate the results, we employed widely recognized NLP metrics to assess the quality of the generated reports in comparison to the reference reports. 
Specifically, we utilized BLEU-1, BLEU-2, BLEU-3, and BLEU-4 to measure the accuracy of n-grams of varying lengths while incorporating a brevity penalty to penalize excessively short predictions~\cite{papineni2002bleu}. 
Additionally, we applied ROUGE metrics, including ROUGE-N to evaluate n-gram recall and ROUGE-L to assess the longest common subsequence, capturing both precision and recall with a focus on sequence-level similarity~\cite{lin2004rouge}. 
Finally, we utilized METEOR, which combines precision and recall through a harmonic mean and incorporates stemming, synonymy matching, and a fragmentation penalty to better align with human judgment~\cite{banerjee2005meteor}. 
These metrics collectively provide a comprehensive evaluation of both lexical and semantic alignment between the generated and reference reports.

\section{Results and Discussion}
\label{results}
Here, we present the outcomes of our investigation into RRG, focusing on model performance in varied linguistic and domain-specific training contexts.
First, we established a robust baseline for English, a well-resourced language, and explored the impacts of domain-specific knowledge by comparing generalist and specialized models. 
Further analyses assessed the effectiveness of instruction tuning, both medical and language-specific, on model performance in low-resource languages. 
Finally, we also examined the role of the temperature parameter in optimizing report generation, providing guidelines for its adjustment.

\begin{figure*}
    \centering
    \includegraphics[width=0.75\textwidth]{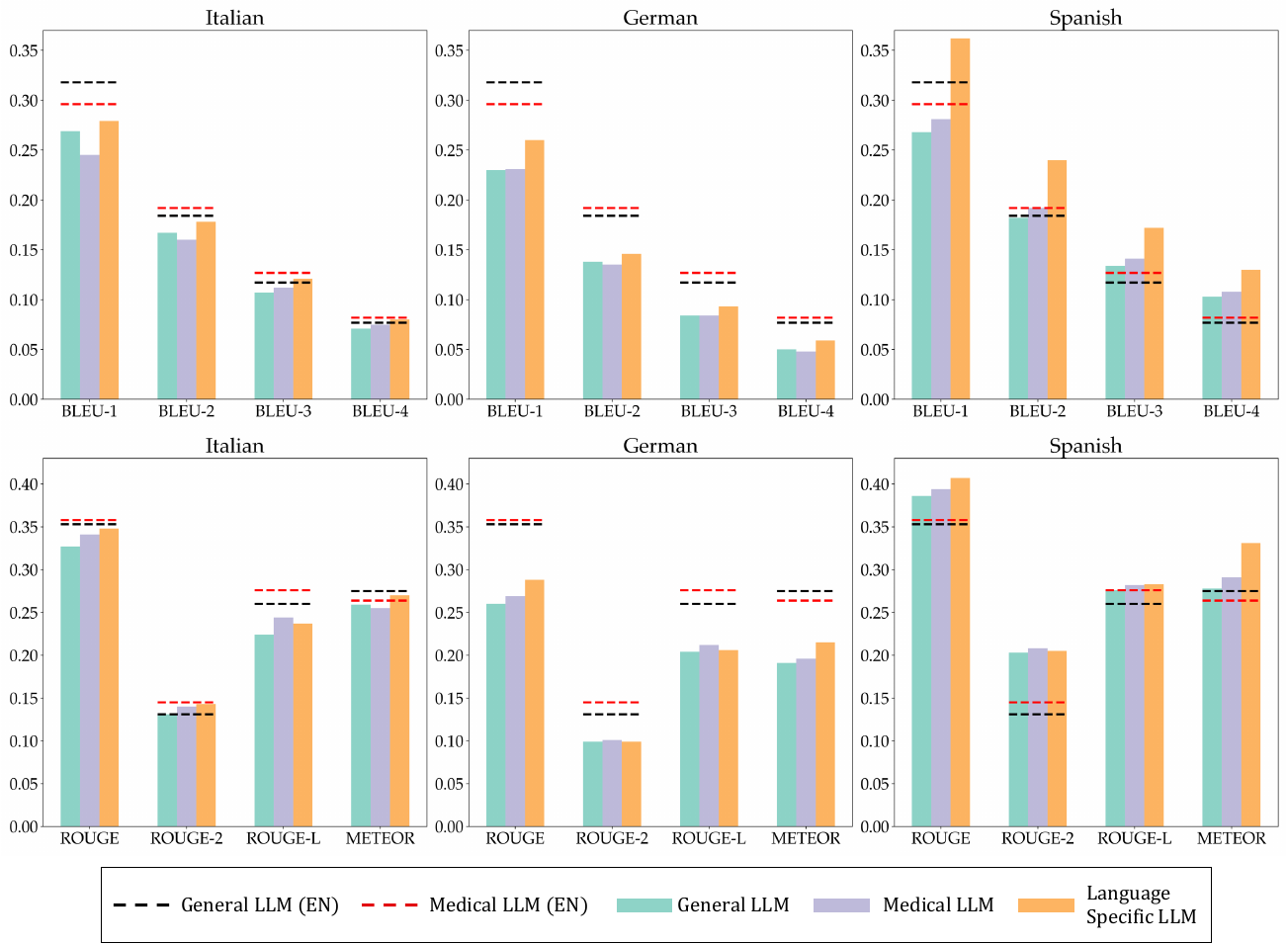}
    \caption{Comparative Histograms of Evaluation Metrics for RRG Models. The top row displays results for the BLEU metric across different models, while the second row presents histograms for the ROUGE and METEOR metrics.}
    \label{fig:3}
\end{figure*}

\subsection{Comparative Evaluation of LLMs} 

Fig. \ref{fig:3} presents a detailed comparison of the performance of General, Medical, and Language-Specific LLMs on the RRG task for each low-resource language analyzed, adding two dashed lines to illustrate the baselines for LLMs fine-tuned directly in English. 
The General LLM fine-tuned on English reports (black dashed line) achieved a BLEU-1 score of 0.318, but its performance declined significantly for longer n-gram sequences, with a BLEU-4 score of 0.077, indicating reduced effectiveness with complex language structures. 
In contrast, the Medical LLM, fine-tuned on the same English reports (red dashed line), achieved a slightly lower BLEU-1 score (0.296) but outperformed the General LLM in BLEU-4 (0.082), reflecting its enhanced ability to handle medical syntax and terminology. 
Furthermore, the Medical LLM outperformed the General LLM across all ROUGE metrics, underscoring its superior semantic and structural alignment with specialized content. 
However, the General LLM recorded a slightly higher METEOR score of 0.275, compared to the Medical LLM's 0.264, reflecting perhaps a broader linguistic adaptability that captures the general semantic content more effectively.
The slight lag in METEOR score for the Medical LLM might be attributed to its focus on medical scientific literature, which, while precise, does not always align with the specific language used in clinical radiology reports.

The adaptation of the General LLM to Italian showed a slight performance decline compared to the baseline, with a BLEU-1 score of 0.269 and BLEU-4 score of 0.071, reflecting challenges in preserving structural complexity. 
The METEOR score of 0.259, though slightly lower than the baseline, still demonstrated reasonable adaptation to the semantic nuances of Italian.
For the Medical LLM, adaptation resulted in lower BLEU-1 (0.245) and BLEU-4 (0.075) scores compared to the English baseline (0.296 and 0.082, respectively), highlighting difficulties in capturing medical terminology in Italian. 
ROUGE (0.341) and METEOR (0.255) scores further confirmed this performance drop.
Differently, with regard the Language-Specific LLM, while its BLEU-1 score was lower than the English baseline, BLEU-4 and METEOR scores were more comparable, emphasizing its strength in maintaining lexical precision and semantic integrity. 
Although its ROUGE score (0.348) was marginally below the baseline, the model's consistent improvement across most metrics stresses its effectiveness in domain-specific adaptation.
In addition, this LLM outperformed the General and Medical LLMs, achieving a BLEU-1 score of 0.279, BLEU-4 score of 0.08, and METEOR score of 0.27, demonstrating superior handling of Italian lexical and semantic nuances. 

In the German context, the General LLM exhibited a marked reduction in performance compared to its counterpart in English, with a BLEU-1 score of 0.23 and BLEU-4 score of 0.05, reflecting challenges in handling German grammar and structure. 
The METEOR score also fell to 0.191, the lowest among the three languages, highlighting substantial challenges in achieving semantic accuracy. 
The Medical LLM showed moderate adaptability, with a BLEU-1 score of 0.231 and a METEOR score of 0.196, indicating limited retention of domain-specific capabilities in German. 
In contrast, the German-specific LLM demonstrated superior performance, achieving a BLEU-1 score of 0.260, the highest METEOR score of 0.215, and consistent results across ROUGE metrics, emphasizing its enhanced ability to capture the linguistic nuances of German radiology reports.

The evaluation of LLMs adapted to Spanish demonstrated competitive performance across various metrics. The General LLM achieved a BLEU-1 score of 0.268, below the English baseline (0.318), but surpassed the English BLEU-4 baseline (0.077) with a score of 0.103, reflecting improved handling of complex language structures. 
Its METEOR score of 0.278 closely aligned with the English baseline (0.275), indicating effective semantic adaptation to Spanish medical language.
The Medical LLM exhibited enhanced performance in Spanish, with BLEU-1 and BLEU-4 scores of 0.281 and 0.108, respectively, surpassing the English BLEU-4 performance.
Notably, the Medical LLM achieved the highest ROUGE (0.394) and METEOR (0.291) scores among medical models, demonstrating superior adaptation to domain-specific terminology in Spanish. 
The Language-Specific LLM recorded the highest overall performance, with a BLEU-1 score of 0.362, a BLEU-4 score of 0.130, a ROUGE score of 0.407, and a METEOR score of 0.331, highlighting its robust adaptation to the structural and semantic complexities of Spanish radiology reports.

The findings from the low-resource language adaptation clearly demonstrate that Language-Specific models consistently outperform both General and Medical models, underscoring the effectiveness of language-specific fine-tuning. 
A detailed analysis of the Italian, German, and Spanish Language-Specific models reveals notable disparities, reflecting the inherent challenges of adapting to specialized domains such as RRG. 
The Spanish model proved to be the most effective, achieving the highest scores across nearly all metrics, including a METEOR score of 0.331, likely due to a more comprehensive pre-training dataset. 
Although the Italian and German models performed well, they lagged behind, suggesting limitations in their pre-training data. 
The Italian model outperformed the German one, especially in managing complex sentence structures, as shown by its superior BLEU-4 score. 
Nonetheless, both underperformed compared to the Spanish model in terms of linguistic fluency and consistency. Overall, this comparative analysis reinforces the value of Language-Specific tuning and highlights the critical importance of well-prepared datasets for improving adaptability to the linguistic and contextual nuances of underrepresented languages.

\begin{figure}
    \centering
    \includegraphics[width=0.48\textwidth]{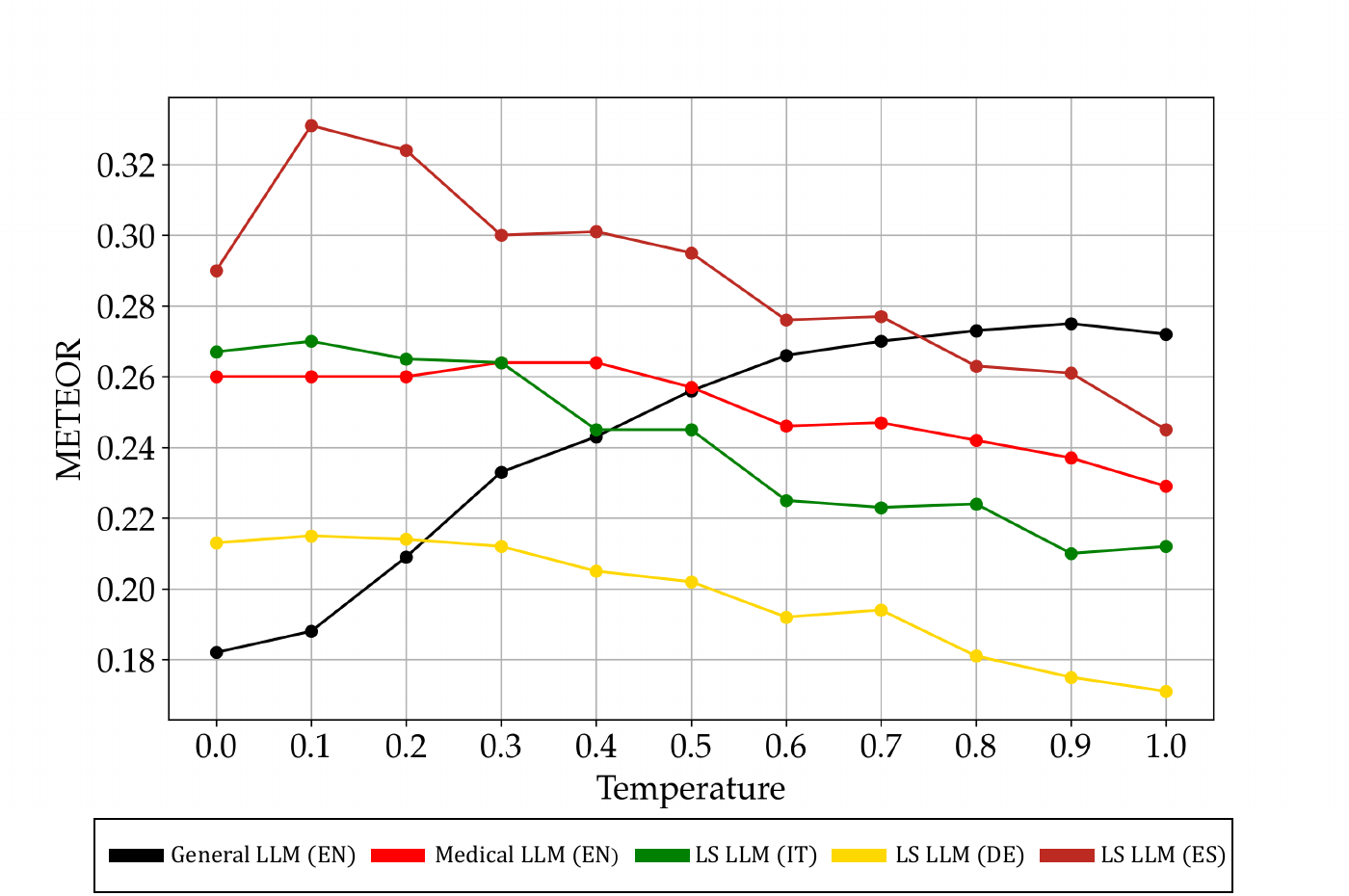}
    \caption{Temperature Analysis.}
    \label{fig:4}
\end{figure}

\subsection{Temperature Analysis}
Finally, we examined the impact of the temperature parameter on METEOR scores, focusing on models with the best performance, such as the General and Medical baseline LLMs in English and the three Language-Specific models (Fig.\ref{fig:4}).
The General LLM demonstrated a trend of gradual improvement, with METEOR scores rising from 0.182 at temperature 0 to a peak of 0.275 at temperature 0.9, followed by a slight decline to 0.272 at temperature 1. 
This trajectory suggests that higher temperatures enhance the model’s capacity to generate diverse expressions, aligning better with the variability characteristic of radiology reports.
In contrast, the Medical model exhibited remarkable stability, starting at a METEOR score of 0.26 at temperature 0, peaking at 0.264 between temperatures 0.3 and 0.4, and decreasing to 0.229 at temperature 1. 
This stability likely stems from the model’s specialized training on medical terminologies, which limits variability but ensures precision.
Language-Specific models revealed distinct behaviors influenced by linguistic and structural features. 
The Italian model achieved its highest METEOR score of 0.27 at temperature 0.1, followed by a gradual decrease to 0.212 at temperature 1. 
Similarly, the Spanish model peaked more prominently at 0.331 at temperature 0.1, highlighting the benefits of slight temperature increases in leveraging creative language variations suitable for radiology reports in Spanish. 
In contrast, the German model showed a more pronounced decline, peaking at 0.215 at temperature 0.1 and decreasing to 0.171 at temperature 1, reflecting the challenge posed by the syntactic complexity of German.
These results indicate that while higher temperatures can enhance creativity and variability, their impact varies depending on the language’s structural and lexical characteristics.

However, consistently, all Language-Specific models achieved their highest METEOR scores at low temperature settings, indicating that minimal output diversity is beneficial for ensuring reliability. These findings underscore the complex interplay between training data size, language model specialization, and temperature configurations. High-resource LLMs, supported by extensive pre-training data, exhibit greater sensitivity to temperature adjustments, as illustrated by the black line in Fig.\ref{fig:4}.
When fine-tuning an LLM on high-resource languages, the abundance of data generally ensures that the model has learned a robust representation of the language, including a wide range of vocabulary, idiomatic expressions, and complex sentence structures. This rich linguistic knowledge base affects how temperature influences performance. At lower temperatures, overfitting becomes a concern, as outputs can become overly deterministic, reflecting an over-reliance on the most frequent patterns observed during training. This reduces the model’s ability to generalize to different contexts or generate creative language. Additionally, lower temperatures can lead to a loss of diversity in the outputs, resulting in repetitive phrasing and limiting expression.
Conversely, Language-Specific LLMs show limited sensitivity to temperature changes, likely due to inherent constraints in output diversity and linguistic complexity. The consistent optimal performance at a temperature of 0.1 highlights the need to balance creativity and clinical accuracy in medical language generation. This setting enables nuanced and varied outputs while maintaining the precision essential for RRG. These dynamics emphasize the importance of careful tuning of the temperature parameter. On the one hand, with limited datasets available, there is a risk of generating overly uniform outputs at low temperatures, compromising the richness necessary for effective report writing. On the other, higher temperatures may introduce hallucinations, jeopardizing clinical accuracy. Striking the right balance is thus crucial to optimize model performance and meet the specific demands of clinical applications, where both linguistic diversity and accuracy are vital.

\section{Conclusions}
\label{conclusions}
This study evaluates the performance of various instruction-tuned VLMs in domain-specific and language-specific contexts, specifically in RRG. 
Our findings indicate that language-specific models excel in generating radiology reports in multiple low-resource languages due to their superior handling of complex linguistic nuances. 
Additionally, models trained with medical terminologies demonstrate enhanced performance across languages, supporting the efficacy of domain-specific tuning in specialized applications. 
The analysis also reveals that the temperature parameter plays a crucial role in balancing coherence and randomness, impacting the overall quality of reports. 
These results underline the importance of language-specific training and suggest that domain-specific models could be improved by tailoring them more closely to the specific medical language of the reports. 

While this study provides valuable insights into the adaptation of VLMs to RRG, several limitations should be noted. 
Firstly, the MedSAM encoder, primarily for image segmentation, was used to assess generalization capabilities of Vision FMs, suggesting that a chest x-ray-specific encoder might improve results. Secondly, the exclusion of report indications from the prompts may have compromised the relevance and quality of the generated reports; future studies could enhance outputs by including these indications. Furthermore, the study's scope was limited to a single dataset and three non-English languages, based on the availability of language-specific models with a Mistral backbone, potentially limiting the generalizability of our findings. Lastly, the absence of established multilingual clinical metrics presents a significant challenge in evaluating clinical effectiveness, highlighting an urgent need for robust clinical metrics to assess multilingual model performance in medical settings.

Our work lays the groundwork for future investigations, aiming to bridge the gap between artificial intelligence potential and its practical utility in improving global healthcare outcomes.

\section*{Acknoweldgments}
Marco Salmè is a Ph.D. student enrolled in the National Ph.D. in Artificial Intelligence, XXXIX cycle, course on Health and Life Sciences, organized by Università Campus Bio-Medico di Roma.
This work was partially founded by: 
i) Università Campus Bio-Medico di Roma under the program ``University Strategic Projects'' within the project ``AI-powered Digital Twin for next-generation lung cancEr cAre (IDEA)''; 
ii) PNRR MUR project PE0000013-FAIR.
iii)  Cancerforskningsfonden Norrland project MP23-1122;
iv) Kempe Foundation project JCSMK24-0094.
Resources are provided by the National Academic Infrastructure for Supercomputing in Sweden (NAISS) and the Swedish National Infrastructure for Computing (SNIC) at Alvis @ C3SE, partially funded by the Swedish Research Council through grant agreements no. 2022-06725 and no. 2018-05973.

\bibliographystyle{IEEEtran}
\bibliography{biblio2}

\begin{thebibliography}{10}
\providecommand{\url}[1]{#1}
\csname url@samestyle\endcsname
\providecommand{\newblock}{\relax}
\providecommand{\bibinfo}[2]{#2}
\providecommand{\BIBentrySTDinterwordspacing}{\spaceskip=0pt\relax}
\providecommand{\BIBentryALTinterwordstretchfactor}{4}
\providecommand{\BIBentryALTinterwordspacing}{\spaceskip=\fontdimen2\font plus
\BIBentryALTinterwordstretchfactor\fontdimen3\font minus \fontdimen4\font\relax}
\providecommand{\BIBforeignlanguage}[2]{{%
\expandafter\ifx\csname l@#1\endcsname\relax
\typeout{** WARNING: IEEEtran.bst: No hyphenation pattern has been}%
\typeout{** loaded for the language `#1'. Using the pattern for}%
\typeout{** the default language instead.}%
\else
\language=\csname l@#1\endcsname
\fi
#2}}
\providecommand{\BIBdecl}{\relax}
\BIBdecl

\bibitem{bommasani2021opportunities}
R.~Bommasani \emph{et~al.}, ``{On the opportunities and risks of foundation models},'' \emph{arXiv preprint arXiv:2108.07258}, 2021.

\bibitem{zhang2024challenges}
S.~Zhang and D.~Metaxas, ``{On the challenges and perspectives of foundation models for medical image analysis},'' \emph{Medical image analysis}, vol.~91, p. 102996, 2024.

\bibitem{van2024large}
M.-H. Van \emph{et~al.}, ``{On large visual language models for medical imaging analysis: An empirical study},'' in \emph{{2024 IEEE/ACM Conference on Connected Health: Applications, Systems and Engineering Technologies (CHASE)}, pages={172--176}}.\hskip 1em plus 0.5em minus 0.4em\relax IEEE, 2024.

\bibitem{guarrasi2024systematic}
V.~Guarrasi \emph{et~al.}, ``A systematic review of intermediate fusion in multimodal deep learning for biomedical applications,'' \emph{Image and Vision Computing}, p. 105509, 2025.

\bibitem{guarrasi2024multimodal}
------, ``Multimodal explainability via latent shift applied to covid-19 stratification,'' \emph{Pattern Recognition}, vol. 156, p. 110825, 2024.

\bibitem{guarrasi2023multi}
------, ``Multi-objective optimization determines when, which and how to fuse deep networks: An application to predict covid-19 outcomes,'' \emph{Computers in Biology and Medicine}, vol. 154, p. 106625, 2023.

\bibitem{rofena2024deep}
A.~Rofena \emph{et~al.}, ``A deep learning approach for virtual contrast enhancement in contrast enhanced spectral mammography,'' \emph{Computerized Medical Imaging and Graphics}, vol. 116, p. 102398, 2024.

\bibitem{ruffini2024multi}
F.~Ruffini \emph{et~al.}, ``Multi-dataset multi-task learning for covid-19 prognosis,'' in \emph{International Conference on Medical Image Computing and Computer-Assisted Intervention}.\hskip 1em plus 0.5em minus 0.4em\relax Springer, 2024, pp. 251--261.

\bibitem{yang2024vision}
C.~Yang \emph{et~al.}, ``{Vision Model Pre-training on Interleaved Image-Text Data via Latent Compression Learning},'' \emph{arXiv preprint arXiv:2406.07543}, 2024.

\bibitem{mishra2021cross}
S.~Mishra \emph{et~al.}, ``{Cross-Task Generalization via Natural Language Crowdsourcing Instructions},'' in \emph{60th Annual Meeting of the Association for Computational Linguistics, ACL 2022}.\hskip 1em plus 0.5em minus 0.4em\relax Association for Computational Linguistics (ACL), 2022, pp. 3470--3487.

\bibitem{wei2021finetuned}
J.~Wei \emph{et~al.}, ``{Finetuned Language Models are Zero-Shot Learners},'' in \emph{International Conference on Learning Representations}, 2021.

\bibitem{sanh2021multitask}
V.~Sanh \emph{et~al.}, ``{Multitask Prompted Training Enables Zero-Shot Task Generalization},'' in \emph{ICLR 2022-Tenth International Conference on Learning Representations}, 2022.

\bibitem{yu2024visual}
B.~X. Yu \emph{et~al.}, ``{Visual tuning},'' \emph{ACM Computing Surveys}, vol.~56, no.~12, pp. 1--38, 2024.

\bibitem{nguyen2024multilingual}
T.~Nguyen \emph{et~al.}, ``{Multilingual Diversity Improves Vision-Language Representations},'' \emph{arXiv preprint arXiv:2405.16915}, 2024.

\bibitem{wang2023adapting}
Y.~Wang \emph{et~al.}, ``{Adapting grounded visual question answering models to low resource languages},'' in \emph{{Proceedings of the IEEE/CVF Conference on Computer Vision and Pattern Recognition}}, 2023, pp. 2596--2605.

\bibitem{zhang2023don}
X.~Zhang \emph{et~al.}, ``{Don't Trust ChatGPT when Your Question is not in English: A Study of Multilingual Abilities and Types of LLMs},'' \emph{arXiv preprint arXiv:2305.16339}, 2023.

\bibitem{magueresse2020low}
A.~Magueresse, V.~Carles, and E.~Heetderks, ``{Low-resource languages: A review of past work and future challenges},'' \emph{arXiv preprint arXiv:2006.07264}, 2020.

\bibitem{muennighoff2022crosslingual}
N.~Muennighoff \emph{et~al.}, ``{Crosslingual Generalization through Multitask Finetuning},'' in \emph{Proceedings of the 61st Annual Meeting of the Association for Computational Linguistics (Volume 1: Long Papers)}, 2023, pp. 15\,991--16\,111.

\bibitem{chen2023monolingual}
P.~Chen \emph{et~al.}, ``{Monolingual or Multilingual Instruction Tuning: Which Makes a Better Alpaca},'' in \emph{Findings of the Association for Computational Linguistics: EACL 2024}, 2024, pp. 1347--1356.

\bibitem{shaham2024multilingual}
U.~Shaham \emph{et~al.}, ``{Multilingual instruction tuning with just a pinch of multilinguality},'' \emph{arXiv preprint arXiv:2401.01854}, 2024.

\bibitem{chen2023meditron}
Z.~Chen \emph{et~al.}, ``{Meditron-70b: Scaling medical pretraining for large language models},'' \emph{arXiv preprint arXiv:2311.16079}, 2023.

\bibitem{li2024llava}
C.~Li \emph{et~al.}, ``{Llava-med: Training a large language-and-vision assistant for biomedicine in one day},'' \emph{Advances in Neural Information Processing Systems}, vol.~36, 2024.

\bibitem{hyland2023maira}
S.~L. Hyland \emph{et~al.}, ``{Maira-1: A specialised large multimodal model for radiology report generation},'' \emph{arXiv preprint arXiv:2311.13668}, 2023.

\bibitem{chaves2024towards}
J.~M.~Z. Chaves \emph{et~al.}, ``{Towards a clinically accessible radiology foundation model: open-access and lightweight, with automated evaluation},'' \emph{arXiv preprint arXiv:2403.08002}, 2024.

\bibitem{chiang2023vicuna}
W.-L. Chiang \emph{et~al.}, ``{Vicuna: An open-source chatbot impressing gpt-4 with 90\%* chatgpt quality},'' \emph{See https://vicuna. lmsys. org (accessed 14 April 2023)}, vol.~2, no.~3, p.~6, 2023.

\bibitem{perez2024rad}
F.~P{\'e}rez-Garc{\'\i}a \emph{et~al.}, ``{RAD-DINO: Exploring Scalable Medical Image Encoders Beyond Text Supervision},'' \emph{arXiv preprint arXiv:2401.10815}, 2024.

\bibitem{yong2022bloom+}
Z.~X. Yong \emph{et~al.}, ``{BLOOM+ 1: Adding Language Support to BLOOM for Zero-Shot Prompting},'' in \emph{Proceedings of the 61st Annual Meeting of the Association for Computational Linguistics (Volume 1: Long Papers)}, 2023, pp. 11\,682--11\,703.

\bibitem{yang2023bigtranslate}
W.~Yang \emph{et~al.}, ``{Bigtranslate: Augmenting large language models with multilingual translation capability over 100 languages},'' \emph{arXiv preprint arXiv:2305.18098}, 2023.

\bibitem{nicolson2023improving}
A.~Nicolson \emph{et~al.}, ``{Improving chest X-ray report generation by leveraging warm starting},'' \emph{Artificial intelligence in medicine}, vol. 144, p. 102633, 2023.

\bibitem{ma2024segment}
J.~Ma \emph{et~al.}, ``{Segment anything in medical images},'' \emph{Nature Communications}, vol.~15, no.~1, p. 654, 2024.

\bibitem{hu2021lora}
E.~J. Hu \emph{et~al.}, ``Lora: Low-rank adaptation of large language models,'' in \emph{International Conference on Learning Representations}, 2021.

\bibitem{jiang2023mistral}
A.~Q. Jiang \emph{et~al.}, ``{Mistral 7B},'' \emph{arXiv preprint arXiv:2310.06825}, 2023.

\bibitem{labrak2024biomistral}
Y.~Labrak \emph{et~al.}, ``{Biomistral: A collection of open-source pretrained large language models for medical domains},'' \emph{arXiv preprint arXiv:2402.10373}, 2024.

\bibitem{mcentyre2001pubmed}
J.~McEntyre and D.~Lipman, ``Pubmed: bridging the information gap,'' \emph{Cmaj}, vol. 164, no.~9, pp. 1317--1319, 2001.

\bibitem{huggingfaceMaestrale}
E.~Federici \emph{et~al.}, ``Maestrale,'' \url{https://huggingface.co/mii-llm/maestrale-chat-v0.4-alpha-sft}, 2024.

\bibitem{leolm2023}
B.~Plüster \emph{et~al.}, ``{LeoLM: Linguistically Enhanced Open Language Model},'' 2023, available at LAION: \url{https://laion.ai/leolm-igniting-german-language-llm-research/}.

\bibitem{huggingfaceOcciglot}
Occiglot, ``Occiglot-es-instruct,'' \url{https://huggingface.co/occiglot/occiglot-7b-es-en-instruct}, 2024.

\bibitem{demner2016preparing}
D.~Demner-Fushman \emph{et~al.}, ``{Preparing a collection of radiology examinations for distribution and retrieval},'' \emph{Journal of the American Medical Informatics Association}, vol.~23, no.~2, pp. 304--310, 2016.

\bibitem{chen2022cross}
Z.~Chen \emph{et~al.}, ``{Cross-modal Memory Networks for Radiology Report Generation},'' in \emph{Proceedings of the 59th Annual Meeting of the Association for Computational Linguistics and the 11th International Joint Conference on Natural Language Processing (Volume 1: Long Papers)}, 2021, pp. 5904--5914.

\bibitem{chen2020generating}
------, ``{Generating Radiology Reports via Memory-driven Transformer},'' in \emph{Proceedings of the 2020 Conference on Empirical Methods in Natural Language Processing (EMNLP)}, 2020, pp. 1439--1449.

\bibitem{liu2021exploring}
F.~Liu \emph{et~al.}, ``{Exploring and distilling posterior and prior knowledge for radiology report generation},'' in \emph{{Proceedings of the IEEE/CVF conference on computer vision and pattern recognition}}, 2021, pp. 13\,753--13\,762.

\bibitem{google_translate}
\BIBentryALTinterwordspacing
``{Google Translate}.'' [Online]. Available: \url{https://cloud.google.com/translate}
\BIBentrySTDinterwordspacing

\bibitem{graves2013generating}
A.~Graves, ``{Generating sequences with recurrent neural networks},'' \emph{arXiv preprint arXiv:1308.0850}, 2013.

\bibitem{papineni2002bleu}
K.~Papineni \emph{et~al.}, ``{Bleu: a method for automatic evaluation of machine translation},'' in \emph{{Proceedings of the 40th annual meeting of the Association for Computational Linguistics}}, 2002, pp. 311--318.

\bibitem{lin2004rouge}
C.-Y. Lin, ``{Rouge: A package for automatic evaluation of summaries},'' in \emph{Text summarization branches out}, 2004, pp. 74--81.

\bibitem{banerjee2005meteor}
S.~Banerjee and A.~Lavie, ``{METEOR: An automatic metric for MT evaluation with improved correlation with human judgments},'' in \emph{{Proceedings of the acl workshop on intrinsic and extrinsic evaluation measures for machine translation and/or summarization}}, 2005, pp. 65--72.

\end{thebibliography}

\end{document}